\documentclass[nonacm]{acmart}

\usepackage{csquotes}
\usepackage[inline]{enumitem}
\usepackage{framed}
\usepackage{booktabs}
\usepackage{graphicx}
\usepackage{subcaption}
\usepackage[utf8]{inputenc}  
\usepackage{longtable}  

\AtBeginDocument{%
  \providecommand\BibTeX{{%
    \normalfont B\kern-0.5em{\scshape i\kern-0.25em b}\kern-0.8em\TeX}}}

\setcopyright{none}
\copyrightyear{2024}
\acmYear{2024}
\acmDOI{XXXXXXX.XXXXXXX}

\acmJournal{DGOV}
\acmVolume{37}
\acmNumber{4}
\acmArticle{111}
\acmMonth{2}

\begin{document}

\title{INACIA: Integrating Large Language Models in Brazilian Audit Courts: Opportunities and Challenges}


\author{Jayr Pereira}
\email{jayr.pereira@neuralmind.ai}
\orcid{0000-0001-5478-438X}
\affiliation{%
  \institution{NeuralMind.ai}
  \city{Campinas}
  \state{São Paulo}
  \country{Brazil}
}
\affiliation{%
  \institution{University of Campinas}
  \city{Campinas}
  \state{São Paulo}
  \country{Brazil}
}

\author{Andre Assumpcao}
\email{aassumpcao@ncsc.org}
\orcid{0000-0003-1669-074X}
\affiliation{%
  \institution{National Center for State Courts (NCSC)}
  \city{Williamsburg}
  \state{Virginia}
  \country{United States}
}
\affiliation{%
  \institution{Brazilian Association of Jurimetrics (ABJ)}
  \city{São Paulo}
  \state{SP}
  \country{Brazil}
}
\author{Julio Trecenti}
\email{jtrecenti@trnv.com.br}
\orcid{0000-0002-1680-6389}
\affiliation{%
  \institution{Terranova Consultoria}
  \city{São Paulo}
  \state{São Paulo}
  \country{Brazil}
}

\author{Luiz Airosa}
\email{airosac@tcu.gov.br}
\affiliation{%
  \institution{Brazilian Federal Court of Accounts (TCU)}
  \city{Brasília}
  \state{Distrito Federal}
  \country{Brazil}
}

\author{Caio Lente}
\email{clente@trnv.com.br}
\affiliation{%
  \institution{Terranova Consultoria}
  \city{São Paulo}
  \state{São Paulo}
  \country{Brazil}
}

\author{Jhonatan Cléto}
\email{jhonatan.cleto@neuralmind.ai}
\affiliation{%
  \institution{NeuralMind.ai}
  \city{Campinas}
  \state{São Paulo}
  \country{Brazil}
}
\affiliation{%
  \institution{University of Campinas}
  \city{Campinas}
  \state{São Paulo}
  \country{Brazil}
}

\author{Guilherme Dobins}
\email{guilherme.dobins@neuralmind.ai}
\affiliation{%
  \institution{NeuralMind.ai}
  \city{Campinas}
  \state{São Paulo}
  \country{Brazil}
}
\affiliation{%
  \institution{University of Campinas}
  \city{Campinas}
  \state{São Paulo}
  \country{Brazil}
}

\author{Rodrigo Nogueira}
\email{rodrigo.nogueira@neuralmind.ai}
\orcid{0000-0002-2600-6035}
\affiliation{%
  \institution{NeuralMind.ai}
  \city{Campinas}
  \state{São Paulo}
  \country{Brazil}
}
\affiliation{%
  \institution{University of Campinas}
  \city{Campinas}
  \state{São Paulo}
  \country{Brazil}
}

\author{Luis Mitchell}
\email{luishr@tcu.gov.br}
\affiliation{%
  \institution{Brazilian Federal Court of Accounts (TCU)}
  \city{Brasília}
  \state{Distrito Federal}
  \country{Brazil}
}

\author{Roberto Lotufo}
\email{roberto@neuralmind.ai}
\orcid{0000-0002-5652-0852}
\affiliation{%
  \institution{NeuralMind.ai}
  \city{Campinas}
  \state{São Paulo}
  \country{Brazil}
}
\affiliation{%
  \institution{University of Campinas}
  \city{Campinas}
  \state{São Paulo}
  \country{Brazil}
}
\renewcommand{\shortauthors}{Pereira, et al.}
\newcommand\hl[1]{\textcolor{blue}{#1}}

\begin{abstract}
  This paper introduces INACIA (\textbf{In}strução \textbf{A}ssistida \textbf{c}om \textbf{I}nteligência \textbf{A}rtificial), a groundbreaking system designed to integrate Large Language Models (LLMs) into the operational framework of Brazilian Federal Court of Accounts (TCU). The system automates various stages of case analysis, including basic information extraction, admissibility examination, \textit{Periculum in mora} and \textit{Fumus boni iuris} analyses, and recommendations generation. Through a series of experiments, we demonstrate INACIA's potential in extracting relevant information from case documents, evaluating its legal plausibility, and formulating propositions for judicial decision-making. Utilizing a validation dataset alongside LLMs, our evaluation methodology presents a novel approach to assessing system performance, correlating highly with human judgment. These results underscore INACIA's potential in complex legal task handling while also acknowledging the current limitations. This study discusses possible improvements and the broader implications of applying AI in legal contexts, suggesting that INACIA represents a significant step towards integrating AI in legal systems globally, albeit with cautious optimism grounded in the empirical findings.
\end{abstract}

\ccsdesc[500]{Applied computing~Law}
\ccsdesc[500]{Information systems~Information extraction}
\ccsdesc[500]{Computing methodologies~Natural language processing}
\ccsdesc[300]{General and reference~Empirical studies}



\maketitle

\section{Introduction}

Large Language Models (LLMs) have emerged as a pivotal innovation in the artificial intelligence space, altering our interaction with digital technologies. Their applications span a diverse spectrum, including but not limited to information retrieval, data extraction, text classification, machine translation, and question answering \cite{KALYAN2023100048}.
Beyond these conventional domains, LLMs demonstrate remarkable capabilities in areas such as standardized testing, creative writing \cite{shanahan2023evaluating}, and even complex decision-making tasks in business and legal environments \cite{sun2023short,chen2023introspective}. This versatility showcases their ability to generate insights and solutions that are contextually relevant. 

In this paper, we introduce a groundbreaking application of LLMs for administrative case processing and assisted decision-making developed jointly by the Brazilian Federal Court of Accounts (TCU, or \textit{Tribunal de Contas da União}, in Portuguese) and experts at several Brazilian research organizations. INACIA (\textit{\textbf{In}strução \textbf{A}ssistida \textbf{c}om \textbf{I}nteligência \textbf{A}rtificial}, in Portuguese, or \textit{AI-Assisted Legal Instruction}) is a system that suggests adjudication directions to court decision-makers -- consisting of structured reasoning around cases' claims, evidence, and relationship to legal provisions in Brazilian law. TCU's staff auditors draw from INACIA's analytical support but ultimately bear the responsibility of making adjudication recommendations to court decision-makers. Its inputs are case filings; outputs are suggestions to the auditor to write the initial instruction\footnote{The initial instruction is the document that contains the directions written by the auditor, including the main facts of the case, the legal basis, and the adjudication proposals.}. INACIA uses a combination of pre-trained large language models (LLMs) and search engines to extract information from case documents and to reason through the extracted information. 

Importantly, INACIA goal's is to expedite case adjudication as it sifts through case documents and connects claims, evidence, and legal provisions. INACIA, however, is not replacing skilled auditors; instead, it is augmenting their productivity. By handling routine and straightforward tasks, INACIA allows auditors to focus on more complex, analytical tasks, thereby deepening the overall effectiveness and efficiency of TCU staff.

The main challenge in developing systems like INACIA is evaluating their performance. How can we ensure the quality of the produced text and extracted data? To address this challenge, we construct a validation dataset from TCU's case filings and propose a methodology that uses LLMs to evaluate the system's outputs. To our knowledge, this is the first evaluation of this kind in the legal domain. We demonstrate that our methodology is highly correlated with human evaluation and can be used to evaluate the system's performance. Our results demonstrate that INACIA can produce high-quality text for the initial instruction but also forewarn the need for further research to improve the quality of the system's outputs.

The remainder of this paper is organized as follows. Section \ref{sec:application_domain} provides an overview of TCU and the role of AI in its operations. Section \ref{sec:related_work} reviews the related work in the field of AI and LLMs in legal settings. Section \ref{sec:inacia} introduces INACIA, detailing its case processing pipeline and the challenges it addresses. Section \ref{sec:experiments} presents the experiments conducted to evaluate INACIA's performance, including the validation dataset and the evaluation methodology. Section \ref{sec:results} discusses the results of the experiments and their implications. Section \ref{sec:discussion} discusses the broader implications of INACIA and the challenges of integrating AI in legal systems. Finally, Section \ref{sec:conclusions} concludes the paper and outlines future research directions.

\section{Application Domain} \label{sec:application_domain}

The Brazilian Federal Court of Accounts is the external control institution that supports the National Congress with the mission of overseeing the budget and the financial execution of the Federal Government. TCU issues rulings that serve to prevent deviations from established norms, offers guidance, recommends enhancements, and, when deemed necessary, imposes administrative sanctions on government agencies and agents implicated in any form of misconduct.

\subsection{Role and Responsibilities}

In most contemporary democracies, a network of institutions exists to regulate, investigate, and potentially sanction individuals occupying elected and non-elected office for improperly using public funds \cite{speck2011auditing}. Despite corruption often dominating media attention, mismanagement of public resources makes up the majority of audit findings across Brazil \cite{ferrazfinan2008}. TCU occupies a premier position within the control network of the Brazilian government and is seen as a model institution for auditing practices across the country.\footnote{Each state within Brazil, along with a select number of municipalities, maintains its audit court, which typically draws inspiration for its guidelines from the Federal Court of Accounts (TCU).}

\subsection{Artificial Intelligence and TCU}

Artificial Intelligence (AI) technologies are not new to TCU. Over the past five years, court developers have worked on systems to classify documents by type, summarize case documents, and efficiently search and retrieve case law information to support decisions. Recently, TCU created its ChatGPT-based conversational assistant to handle unstructured, internal information. In 2023, TCU inaugurated an Artificial Intelligence Center (or \textit{Núcleo de Inteligência Artificial}, in Portuguese), whose goal is to guide and implement responsible, ethical use of AI in the Court.

\subsection{Audit Case Processing}

TCU investigates (and eventually issues rulings on) the use of federal funds in two ways: 
\begin{enumerate*}
    \item when TCU independently audits federal agencies (or anyone using federal funds) or
    \item when TCU receives complaints filed by individuals, private entities, or public agencies on the misuse of federal funds.
\end{enumerate*} 

While case proceedings may vary based on the initiation method, case processing is typically a time-intensive task primarily handled by TCU staff. Auditors must review case filings, confirm whether TCU should hear the case, search and append external evidence of wrongdoing (if necessary), check claims against case law and federal statutes, and apply the Court's jurisprudence.
Finally, they prepare a document called \textit{Instrução} (Portuguese for Instruction) that summarizes the case, the evidence, and the legal analysis and proposes adjudication recommendations. The instruction is then sent to the court's judicial panel for final adjudication.
Despite the variability in human effort required for each case, it can be reasonably estimated that the review of a single case necessitates approximately 30 hours, incurring a cost of approximately \$1,750 before the case is even presented to the court's judicial panel. Given an annual average processing rate of about 2,000 cases within the INACIA project's purview and the task's textual nature, audit case processing is an ideal candidate for an LLM application that would help expedite review while reducing costs.\footnote{For comparison purposes, INACIA processes a case in under 20 minutes and at an initial cost of \$10, subject to further human review. A mere 1 percent reduction in the time required for a human to process a case would suffice to offset the costs associated with INACIA.}

\subsection{Large Language Models in Brazilian Legal Settings}
Though natural language processing (NLP) applications have been familiar to courts both in Brazil and worldwide for some time, the use of large language models as the engine behind NLP applications is relatively new -- only recently courts across the world have begun using LLMs for document summarization, chatbots, writing briefs, etc. \cite{butler2024generative}

TCU is a particularly interesting use case as it predates the popularization of LLMs through OpenAI GPT models in late 2022 and throughout 2023. Acknowledging the rapid evolution of AI and in line with its mission to foster innovation and enhance public administration, TCU carried out a series of pre-commercial procurement (PCP) studies between 2019 and 2022 to design a public procurement call for an AI-powered solution to expedite audit case processing. Open for bidding in early 2022, this procurement call was the first of its kind in Brazil as it sought to contract a pre-market AI product requiring research and development and carrying significant technological risk (the solution envisioned by TCU should go beyond the state-of-the-art NLP tools available then).

INACIA was the winning proposal of this procurement call. A team of researchers at the State University of Campinas (UNICAMP) and start-ups NeuralMind and Terranova Consultoria put together the only proposal to use an LLM to handle all NLP tasks needed in the audit processing product (detailed in the following sections). Back in early 2022, LLMs were fairly unknown outside of the NLP expert community and were not consensually seen as the way forward in NLP research. TCU, however, trusted the INACIA proposal, and this manuscript is based on the work done during the proof of concept (POC) development stage.

TCU goals are to focus auditors' expertise on the most challenging tasks of case processing while handing over the more routine tasks to the AI system. This strategy is well-aligned with the evidence on how best to incorporate technology for productivity gains and labor output augmentation \cite{ACEMOGLU20111043}. The current expectation is that AI will change how TCU handles cases from a sequential, step-by-step approach to a more holistic approach in which filings are evaluated simultaneously and connected to external resources necessary for decision-making from the outset. Effectively, this system would eliminate redundancies and streamline case processing without requiring regulation change.

\section{Related Work} \label{sec:related_work}

Artificial Intelligence (AI) and Large Language Models (LLMs) are now part of products and services across various jurisdictions and domains, highlighting a global movement towards technologically enhanced legal systems. Some of these include \citet{chalkidis2020legalbert}, who developed Legal-BERT, a domain-specific adaptation of BERT for legal documents, demonstrating the effectiveness of language models in legal document analysis and information extraction. Similarly, \citet{safaei2023theendofpolicy} focused on zero-shot text matching for financial auditing using LLMs, showcasing the potential of AI in improving the accuracy and efficiency of legal and financial document processing. The work of \citet{Hillebrand2023improving} further extends this by presenting ZeroShotALI, a system that leverages LLMs for recommending relevant legal texts, underlining the advancements in AI-driven legal recommendation systems.

Additionally, the exploration of GPT models in zero-shot settings for legal text annotation by \citet{Savelka_2023} provides insights into the capabilities of LLMs in the semantic understanding and classification of legal documents. \citet{paul2023pretrained} complements this investigation into pre-training LLMs on Indian legal corpus, which enriches the models' understanding of legal texts from different legal systems and highlights the adaptability and scalability of AI techniques in diverse legal environments.

Finally, \citet{sipra2024byte} assesses the challenges and opportunities in AI-driven criminal justice systems, addressing ethical, transparency, and accountability issues, which are critical in the application of AI within legal systems. These works collectively underscore the transformative potential of AI and LLMs in legal research, practice, and administration, offering valuable insights for the integration of such technologies into the Brazilian Federal Court of Accounts (TCU) through INACIA.

\section{The INACIA Project} \label{sec:inacia}

\subsection{Overview}

As stated previously, INACIA's goal is to assist TCU auditors in streamlining case adjudication while refocusing their tasks to high-value-added work, e.g., replacing routine tasks of information extraction for tasks requiring critical reasoning over the combination of a case's claims, the applicable legal framework and evidence brought forth by plaintiffs. The main advantage of using pre-trained LLMs is that they can work in a few-shot or zero-shot setting, meaning that they can be used to process cases without the need for training on a specific dataset or domain. This is particularly important in the TCU context, as the number of cases is relatively small compared to the datasets usually used to train LLMs. OpenAI's models, for example, are trained on a huge dataset of internet text,\footnote{OpenAI does not disclose the exact size of the dataset used to train its models, but it is known to be in the order of hundreds of gigabytes.} which allows them to understand and generate text in a wide range of domains and languages. Using search engines, on the other hand, allows INACIA to access external information, such as case law and legal provisions, which are not present in the case documents and potentially not present (or present in an insignificant amount) in the training data of the LLMs. It allows INACIA to overcome one of the most common limitations of LLMs, which is the lack of access to external information and the tendency to generate text that is not grounded in reality (a.k.a. hallucination). 

\begin{figure}[ht]
  \centering 
  \includegraphics[width=\textwidth]{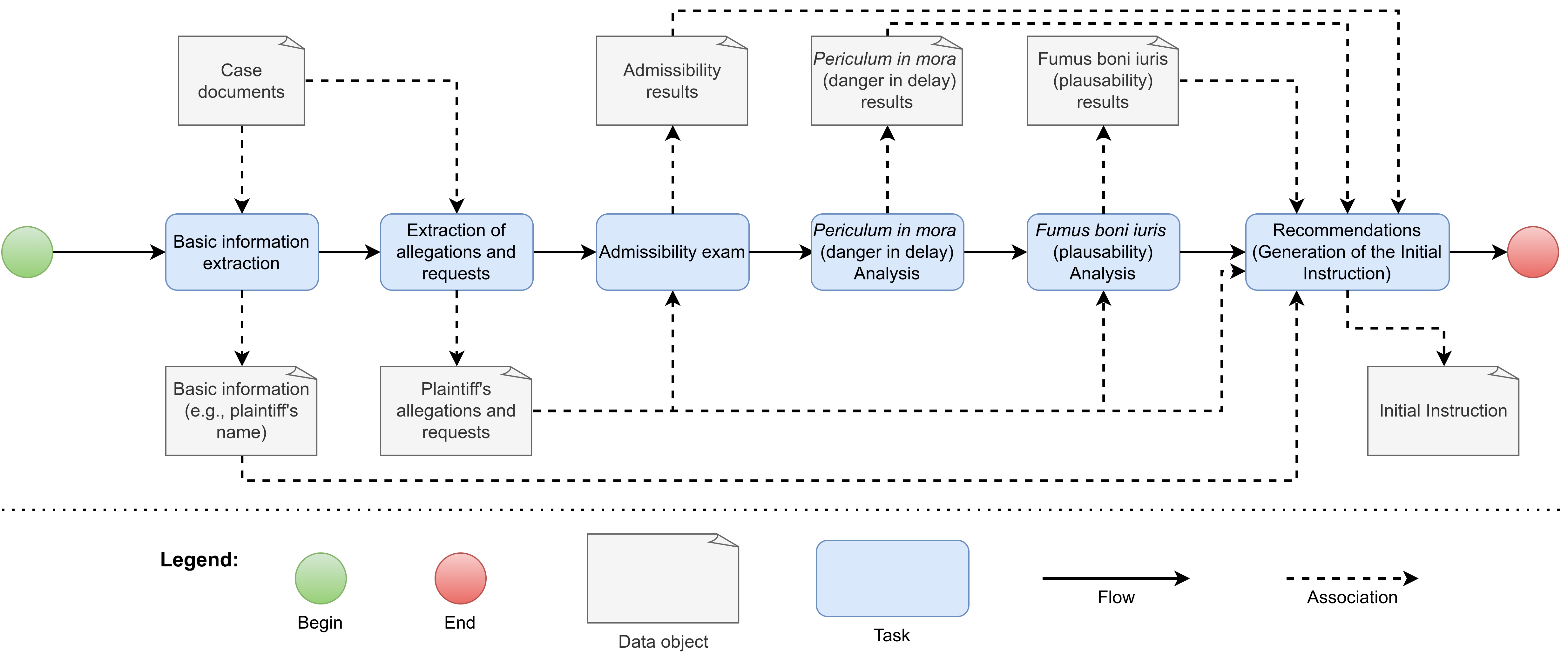}
  \caption{End-to-end INACIA pipeline (from case documents to initial instruction).}
  \label{fig:pipeline}
\end{figure}

Figure \ref{fig:pipeline} depicts INACIA's main case processing pipeline, showcasing the holistic nature of case analysis and how much of the work done by auditors can be replaced by INACIA work. While there is a direct, sequential flow (i.e., solid arrows) of tasks (i.e., blue boxes) needed to produce the instruction (case adjudication suggestion), the dashed lines show the association (or dependency) between tasks. The lighter grey boxes indicate data objects needed to start processing or to be produced as intermediate or final products for other TCU systems.\footnote{
INACIA is available to TCU in two formats: \begin{enumerate*}
    \item a Python API that can be executed end-to-end from an interactive shell, e.g., a Jupyter Notebook, or from source or via classes and methods that handle each component of the pipeline as an independent module; and
    \item a staging environment developed for presenting INACIA's proof of concept to the business team at TCU.
\end{enumerate*} 
In either case, the outputs of one INACIA module are the inputs to the next module. INACIA is currently designed to run end-to-end with no human intervention -- though its execution could be broken out and interspersed with manual checkpoints along the way. The backend of INACIA is entirely structured in the Azure Cloud and the user has the option of choosing any of the pre-trained language models available by Microsoft on its Azure Portal.
}
In the next subsections, we describe each element of this pipeline in detail.

\subsection{Inputs}

The pipeline starts off ingesting all documents available at case submission (\textquote{Case documents} grey box). There are two sets of documents in the submission package: \begin{enumerate*}
    \item the main document is the \textit{representação} (in Portuguese), which is in nature similar to a civil complaint. It is a formal document laying out the claims of wrongdoing and why TCU should take the case;
    and \item supporting documents containing the evidence of wrongdoing as presented by the plaintiff. For example, if the plaintiff alleges that the defendant has not complied with a contract, the contract is a document that supports the case.
\end{enumerate*} 

\begin{figure}[ht]
    \begin{subfigure}{.5\textwidth}
        \includegraphics[width=\linewidth]{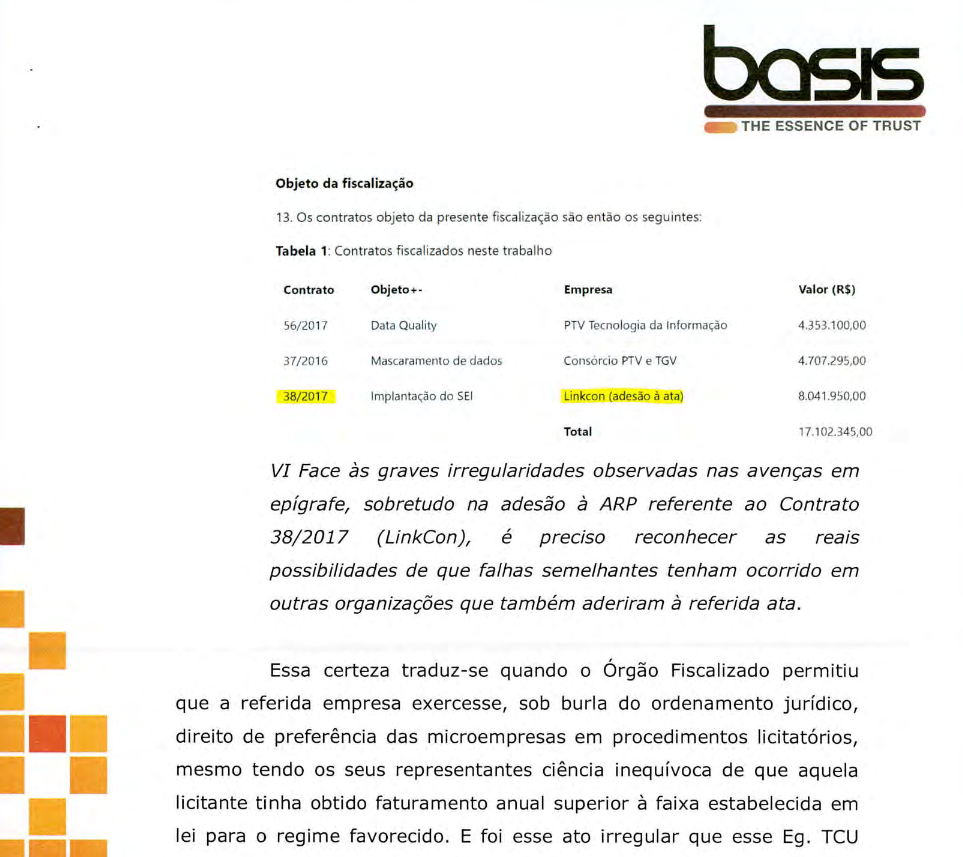}
        \caption{A \textit{representação}}
        \label{fig:image1}
    \end{subfigure}
    \begin{subfigure}{.5\textwidth}
        \includegraphics[width=\linewidth]{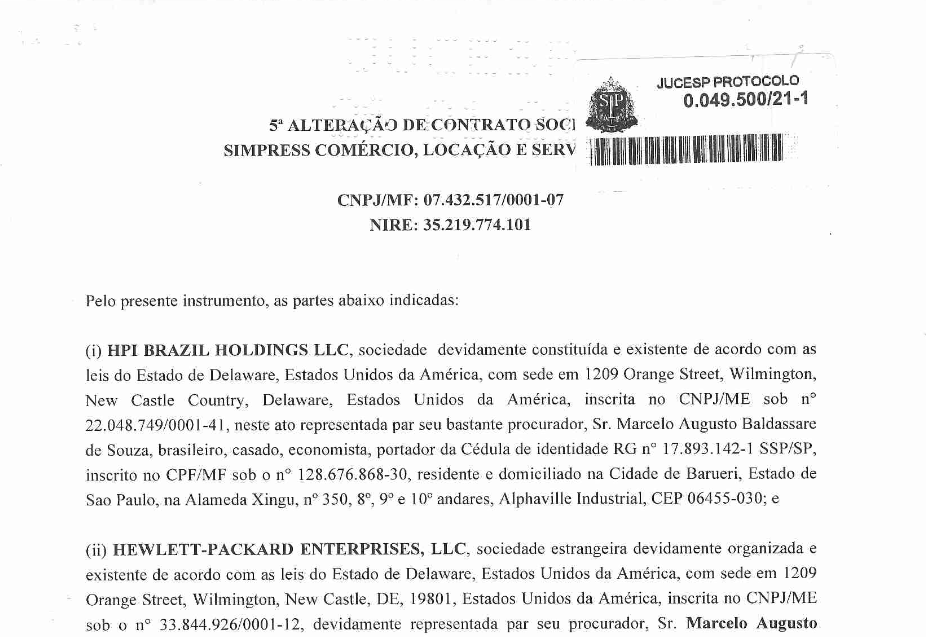}
        \caption{A supporting document (business incorporation documents)}
        \label{fig:image2}
    \end{subfigure}
    \caption{Examples of files typically found in TCU's case database.}
    \label{fig:side-by-side}
\end{figure}

None of these documents follow a strict format. Lawyers are free to petition TCU in any way as long as the main sections of their complaint follow the guidelines set by the TCU. The \textit{representação}, thus, is a free-form text document and the supporting documents can be anything (from pictures to invoices to full procurement contracts). There is no set range of pages for either type of document. 
Figure \ref{fig:side-by-side} contains snapshots of files typically found in TCU's case database.
Figure \ref{fig:image1} shows a \textit{representação} in a native PDF format, and Figure \ref{fig:image2} shows a supporting document (a business incorporation document) in a scanned image format.

Both the \textit{representação} and supporting documents are stored in multiple formats, and they can be classified in one of two categories: \begin{enumerate*}
    \item \textbf{searchable files:} electronically-created files with searchable content (e.g., native PDF document or MS Word document printed as PDF);
    \item \textbf{unsearchable files:} scanned physical files or images whose content is not searchable (e.g., a photo of a contract or signature).
\end{enumerate*} Since INACIA uses the textual content of case documents to perform its tasks, we needed to implement data extraction solutions to handle all file types. These solutions needed to deliver high-quality extraction at a low cost for the solution to be sustainable once TCU takes the system to production. 

Most files in TCU's case database are searchable and primarily stored as PDF documents. Thus, the first implemented solution is a standard PDF extraction pipeline using the Apache Tika framework \cite{apachetika}, which can handle all types of searchable files in TCU's case database. Besides being an open-source tool (thus meeting the low-cost criterion), Tika delivered better results in text extraction than PyPDF2 \cite{pypdf2}, another popular framework for processing searchable PDF documents. We compared and contrasted Tika and PyPDF2 in the following ways:
\begin{enumerate*}
    \item which framework was better at extracting textual content and its structure (e.g., a two-column document processed as separate, two columns instead of a single column with left-to-right text mixing up the two columns).
    \item which framework was better at extracting characters from different encoding formats (e.g., those not present in UTF-8, Latin-1, or Windows-1252 encodings).
\end{enumerate*}
Tika performed better in both criteria -- we particularly liked that it reports the count of invalid characters in the processed text as extraction metadata, allowing us to create a measure of extraction quality by  computing the percentage of invalid characters over all characters extracted.

Whenever Tika fails in text extraction (i.e., it reported zero searchable characters after processing the text), INACIA then deploys the second text extraction solution, which relies on Microsoft's Azure Form Recognizer service for Optical Character Recognition (OCR) \cite{microsoftazureformrecognizer}. Form Recognizer ingests any unsearchable file and returns the extracted text in plain format. We also compared Form Recognizer to other industry-leading OCR frameworks (Tesseract \cite{smith2007overview}, and Google Cloud Vision \cite{googlecloudvision}) using the high-quality and low-cost criteria. While they are similarly priced, Form Recognizer had higher-quality results and richer extraction metadata (e.g., information on the position and content of tables).
To assess text extraction from either document type, we evaluated results qualitatively. We manually compared extracted elements from case documents, such as the position of tables, columns, and text, allowing for a more nuanced understanding of the strengths and weaknesses of each tool in practical scenarios. We evaluated a sample of documents from TCU's database containing all cases and their documents between 2019 and 2022. We pulled out a minimum of five documents per year. Within these documents, we had at least one document of easy, medium, and hard extraction, determined by the number of pages and their content as follows:
\begin{enumerate*}
    \item Easy: up to 10-page documents with minimal tables;
    \item Medium: up to 25-page documents with structured data or images; and
    \item Hard: over 25-page documents with structured and unstructured data elements, images, or handwritten content.
\end{enumerate*}

\subsection{Extraction of Basic Information and Allegations and Requests} \label{sec:basic_info_extraction}

The first two tasks (blue boxes in Figure \ref{fig:pipeline}) are the extraction of case summary information (\textquote{Basic information extraction} blue box), the allegations of wrongdoing made by the plaintiff (allegations), and what action they ask TCU to take (requests) (\textquote{Extraction of allegations and requests} blue box). We describe each task in the sections below.

\subsubsection{Basic Information Extraction}

\begin{figure}[ht]
  \centering 
  \includegraphics[width=0.8\textwidth]{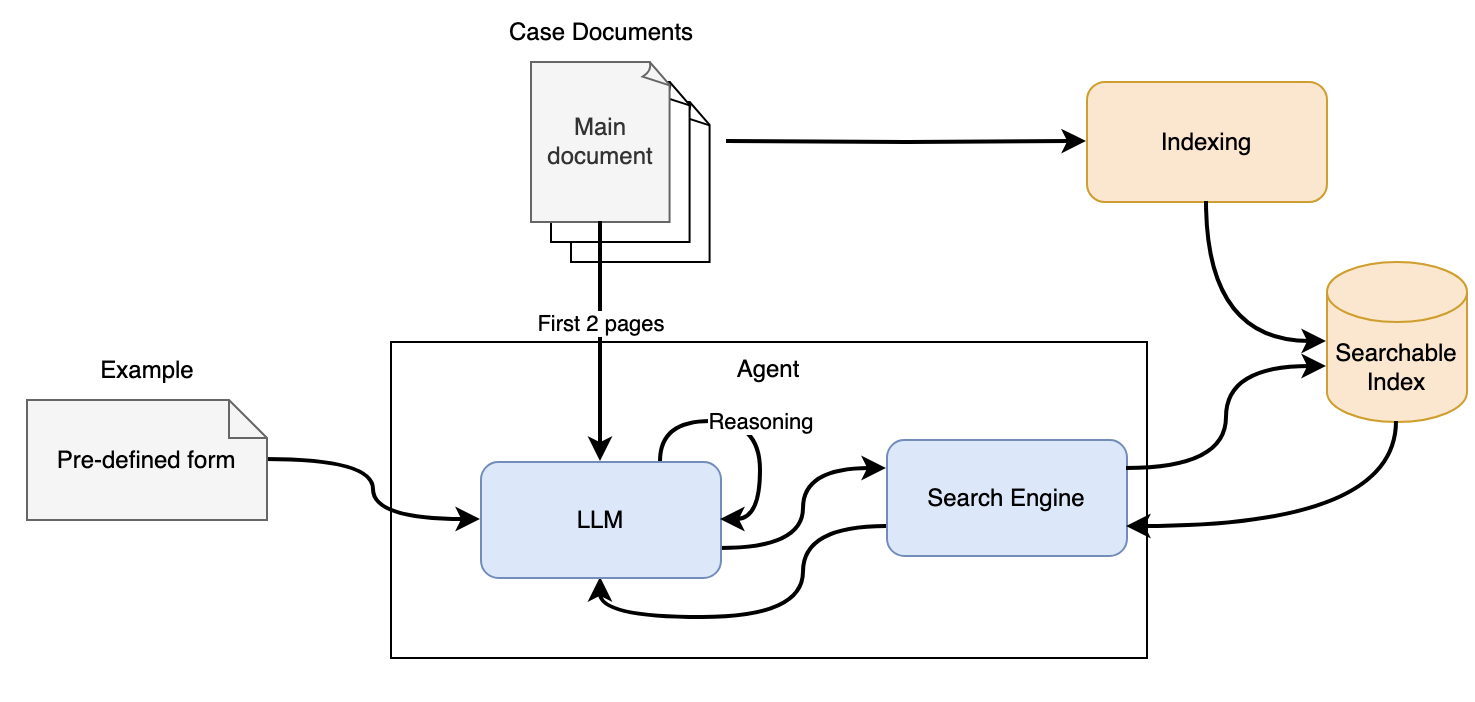}
  \caption{Pipeline used for Basic Information Extraction.}
  \label{fig:basic_information_extraction}
\end{figure}

Figure \ref{fig:basic_information_extraction} illustrates the approach used for basic information extraction. The solid lines indicate the flow of information between components of the system (in grey are the data objects, in blue the agent components; in orange technologies external to the main INACIA pipeline); the work done by INACIA's basic information extraction agent is depicted by the bounded box. It is a structured and systematic process relying on prompt engineering and retrieval-augmented generation (RAG) techniques \cite{lewis2020retrieval}. First, we feed the LLM a structured list of the information we need to extract from case documents (\textquote{Pre-defined form} green box). This list was constructed using older cases and their \textit{representação} containing the information TCU needs to retain for record-keeping purposes.\footnote{
\textquote{Basic information} examples are the case ID, case type, plaintiff name, plaintiff ID, etc.} We also provide the model with detailed explanations of what each of the variables is. Finally, we append to our combined prompt from above the first two pages of a case's \textit{representação} (\textquote{Main document} gray box), which should contain most of the information needed.

We deploy our RAG workflow when we cannot find all necessary information in the first pages of the \textit{representação}. Along with extracting all textual content from case documents, we created searchable indexes of case documents. We allowed the LLM to reason through the form and query such indexes if it did not find the information in the first two pages of the main case document. An example that helps illustrate this mechanism is that TCU staff often need to look at federal agencies' procurement documents to evaluate whether the agencies have followed all necessary rules when contracting external vendors. INACIA can search supporting documents of a case to find procurement contracts and retrieve the necessary information (the rightmost side of Figure \ref{fig:basic_information_extraction} illustrates this process). We use standard industry-leading search algorithms, such as BM25 \cite{Robertson1994OkapiAT}, to retrieve the relevant passages from the indexed case documents. The LLM then processes these passages to extract answers, ensuring a streamlined and consistent approach. For retrieving the remaining information using BM25, we manually crafted pre-defined queries. The resulting information is then fed into the LLM for the final extraction, completing the form. In this workflow, the LLM functions not as an intelligent agent but rather as a tool within the pipeline for interpreting search results.

\subsubsection{Extraction of Allegations and Requests} \label{sec:allegations_requests}

An important step in our pipeline is to assess whether a case's claims and supporting evidence are strong. Often, TCU staff can turn down the review of a case when there is weak evidence of wrongdoing, when the plaintiff cannot straightforwardly articulate their claims that the defendant has broken the law in some shape or form, or even when the plaintiff does not clearly states their request for TCU action. As with other tasks, extracting allegations and actions requested by plaintiffs is a time-consuming task suited for LLM processing.

\begin{figure}[ht]
    \begin{framed}
    \begin{flushleft}
        \fontsize{7pt}{8pt}\selectfont
        \textcolor{teal}{\#\# Instructions}
        \newline
        \newline
        You are an intelligent agent capable of reasoning and interpreting legal documents. You are given a case main document and an example of the allegations and requests to be extracted from the case main document. You must extract the allegations and requests from the case's main document.
        \newline
        \newline
        \textcolor{teal}{\#\# Example}
        \newline
        \newline
        You must provide a list of allegations and requests similar to the example provided below:
        \newline
        \newline
        \textcolor{violet}{[EXAMPLE TEXT GOES HERE.]}
        \newline
        \newline
        \textcolor{teal}{\#\# Rules}
        \newline
        \newline
        You must follow the rules below:
        \newline
        1. Identify the allegations and requests presented by the plaintiff. They can be related to various aspects, including, but not limited to, the violation of a law, regulation, or contract.
        \newline
        2. Enumerate the allegations and requests in the same order as they appear in the case main document.
        \newline
        \textcolor{gray}{...}
        \newline
        \newline
        \textcolor{violet}{[CASE MAIN DOCUMENT TEXT GOES HERE.]}
    \end{flushleft}
    \end{framed}
    \caption{Prompt used to extract the allegations and requests in a \textit{representação}.}
    \label{fig:allegations_prompt}
\end{figure}

In Figure \ref{fig:allegations_prompt}, we present the prompt used at this pipeline step. It has four main characteristics: 
\begin{enumerate*}
    \item the instruction (identified by the title \textquote{\#\# Instruction}), which tells the model its role and what we want it to do;
    \item the example (identified by the title \textquote{\#\# Example}), which teach the model the expected structure and language of the text;
    \item the rules (identified by the title \textquote{\#\# Rules}), which are the constraints that the model must follow to generate the extract the allegations and requests; and
    \item the input, i.e., the case's main document,  from which the model will extract the allegations and requests.
\end{enumerate*}

For this step, we used OpenAI's GPT-4 model with a context length of 32,000 tokens. The main advantage of this model is that it does not require fine-tuning or additional training. Besides, it was trained to generate human-like text, which is essential for the INACIA project. The high context length is also important because the documents can be long and complex, and the model needs to understand the context to generate the allegations and requests correctly.

The data extracted in this step (allegations and requests) are used in three ways: \begin{enumerate*}
    \item for the admissibility examination, where INACIA reviews cases for elements that would make such case \textquote{of public interest} and whether the claims are strong enough to hear the case;
    \item for drafting the initial instruction document (cf. Section \ref{sec:recommendations});
    \item for the \textit{Fumus Boni Iuris} analysis, which means evaluating a case for the likelihood of success on its merits (cf. Section \ref{sec:plausability}).
\end{enumerate*}

\subsection{Admissibility Examination} \label{admissibility}

The next step in the INACIA project is to conduct an admissibility examination (\textquote{Admissibility exam} blue box in Figure \ref{fig:pipeline}). The admissibility examination is a preliminary examination conducted by auditors to determine whether the claims meet the minimum legal criteria to proceed (i.e., whether TCU should hear the case). TCU defines five criteria that must be met for a claim to be admissible: \begin{enumerate}
    \item \textbf{Legitimacy:} the case should be submitted by a legitimate plaintiff (as defined in TCU's bylaws);
    \item \textbf{Competency:} whether the wrongdoing falls within the jurisdiction of TCU (\textquote{competency} is the literal translation from Portuguese);
    \item \textbf{Existence of evidence:} whether there is enough evidence supporting plaintiff's claims;
    \item \textbf{Existence of public interest:} whether the case is of public interest (as determined by potential (non-)monetary damage to the federal government).
    \item \textbf{Clear writing and language:} whether the \textit{representação} has been written clearly and objectively.
\end{enumerate}

\begin{figure}[b]
  \centering 
  \includegraphics[width=\textwidth]{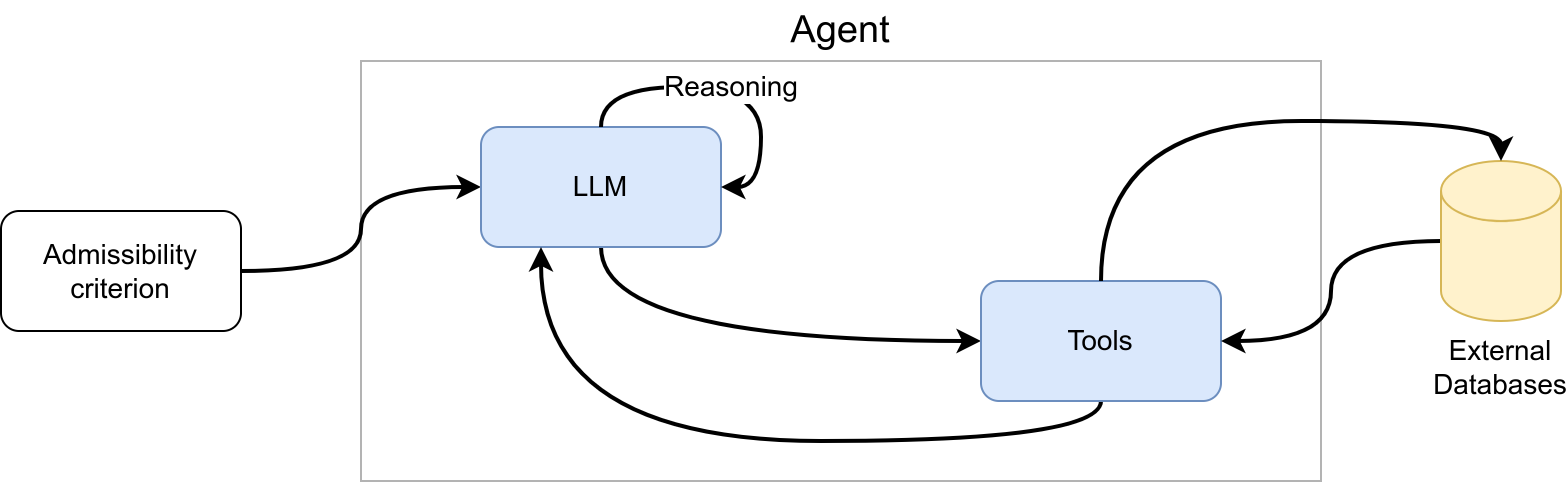}
  \caption{Approach used for Admissibility Examination.}
  \label{fig:admissibilidade}
\end{figure}

As shown in Figure \ref{fig:admissibilidade}, we used an approach similar to the one used for basic information extraction. We applied an LLM as an intelligent agent capable of reasoning and interpreting legal documents. However, instead of feeding the LLM with the first two pages of the case's \textit{representação}, we allow the model to search for information in case documents using a search engine -- including four external databases: \begin{enumerate*}
    \item TCU's jurisprudence database;
    \item TCU's statutes and Federal law;
    \item TCU's internal codes and other documents; and
    \item the case's documents.
\end{enumerate*}

The admissibility examination is initiated by feeding the LLM one admissibility criterion. Then, the LLM searches for information in case documents and external databases to determine whether the case meets such admissibility criterion. For example, consider the admissibility criterion of legitimacy. The agent may search for the plaintiff's information in the case documents and for information about who can file a claim in the external databases (e.g., the court's internal codes). If the agent finds information that confirms a plaintiff's legitimacy, the agent concludes that the claim meets the admissibility criterion. Otherwise, the agent concludes that the claim does not meet the admissibility criterion. This process is repeated for each admissibility criterion. 

In this task, the agent can compose queries to search for information in case documents and external databases. For example, the agent may want to compose the query \textit{quem tem legitimidade para apresentar uma representação ao TCU?} (i.e., who has legitimacy to file a claim to the TCU?) to search for information on who can file a claim in external databases. If that's the case, the agent should not be limited to pre-defined queries and keyword-based queries. 
To facilitate this, we employ NeuralSearchX \cite{almeida2022neuralsearchx}, a search engine that integrates the capabilities of the BM25 algorithm and a neural reranker \cite{nogueira2020document}, fine-tuned for Brazilian Portuguese. NeuralSearchX first retrieves documents containing the words from the query and then utilizes its re-ranking mechanism to prioritize the most relevant documents. 

In addition to searching, we used an approach based on the Visconde system \cite{visconde} to reason through the information retrieved from case documents and external databases. Visconde is a multi-document question-answering system integrating information retrieval algorithms and LLMs to answer questions. The reasoning step is based on the Chain-of-Thought approach \cite{kojima2022llm} and produces a paragraph explaining how the documents support the model's answer.

\subsection{\textit{Periculum in mora} Analysis}

The \textit{Periculum in Mora} (Latin for \textquote{\textit{danger in delay}}) analysis is an essential part of INACIA, as it relates to the urgency with which TCU needs to take action in any particular case. It determines whether immediate action is required to prevent further damage or prejudice to the parties involved.

For this analysis, we carry out a search for exact matches of \textit{minuta de contrato} (draft contract) in all case documents after a case has been assigned to a TCU staff, except when the case contains fewer than five documents -- in which case all documents are considered. This search is essential to flag documents mentioning the draft contract so they are not included in subsequent steps.
Two key rules are used to standardize the decision on whether there is \textit{Periculum in mora}: \begin{enumerate*}
    \item the existence of an active or signed contract -- there is no danger in delay since TCU cannot prevent a contract from coming into force; 
    and \item the existence of events that prove delay, cancellation, or impediment in contracting -- there is no danger in delay because the execution of a contract has been halted.
\end{enumerate*}  The verification of these rules occurs step-wise. We first verify whether there is any indication that a goods or service provision contract has been signed or is active. The verification is done directly from a keyword search and indirectly by asking the LLM to reason through the documents and verify the existence and status of procurement contracts. The second step is verifying any event delaying, canceling, or suspending a contract from entering into force. Here, we do not perform a direct keyword search for such information; we send documents and prompts straight to the LLM. The outcome of these searches provides an answer regarding the existence of an active contract and any delays or impediments.
Finally, the findings are synthesized into a dictionary containing information about the presence of an active contract or delays, along with the document IDs from each search and a suggested decision on whether there is \textit{Periculum in mora}, which can be either \textquote{rejected} or \textquote{accepted}.

It is important to note that the role of the LLM is limited to receiving search results and determining the existence of an active contract or impediments and delays. 
The decision-making flows are hard-coded and not subject to the LLM's discretion. After the decision is made, the result is passed onto the LLM along with a prompt laying out decision rules, enabling the LLM to draft the text that will be included in the initial instruction document. This process ensures that the conclusions drawn from the \textit{Periculum in mora} analysis are accurately reflected in the official documentation, preserving the integrity of TCU's case review process.

\subsection{\textit{Fumus boni iuris} Analysis} \label{sec:plausability}

The \textit{Fumus boni iuris} analysis (Latin for \textquote{\textit{smoke of good law}}) is also an essential part of INACIA; it refers to the \textquote{appearance of good law} or the likelihood of success on the merits of the case. This principle is applied to assess whether the plaintiff's claims appear to be grounded in federal law and to have a reasonable chance of success should the case go to trial. INACIA uses its LLM to analyze the substance of the allegations and the legal principles they invoke. This involves thoroughly examining the legal arguments, jurisprudence, and relevant statutory laws. The LLM assesses the coherence of the allegations with existing legal norms and precedents, providing an initial legal opinion on the case's merit.

For the \textit{Fumus boni iuris} analysis, we used an approach similar to the one used for the admissibility examination. We used the LLM as an intelligent agent capable of reasoning and interpreting legal documents. However, instead of feeding the LLM with one of the admissibility criteria, we feed the LLM with the allegations extracted in the pipeline task \textquote{allegations and requests} (cf. Section \ref{sec:allegations_requests}). The LLM then searches for information in case documents and external databases to determine whether there is \textit{Fumus boni iuris}.

The process for the \textit{Fumus boni iuris} analysis is divided into two steps. In the first step, we feed the agent the allegations extracted in the previous project stage and ask him to classify each allegation as grounded in law, not grounded in law, or inconclusive. 
We use the reasoning and acting components of the ReAct framework \cite{yao2023react} to instruct the model to classify the allegations. The reasoning component is used to reason through the allegations and plan the actions to be taken. The acting component executes the actions planned by the reasoning component, which includes searching the case documents and the external databases. One of the external databases used in this project stage is TCU's selection of statutes and Federal law. The model searches for information about the laws invoked by the allegations in this database. The agent also searches for information about the allegations in the TCU's jurisprudence database. The model then classifies the allegations as grounded in law, not grounded in law, or inconclusive.

The second step consists of producing a decision about the \textit{Fumus boni iuris} of the case. The decision is based on the classification of the allegations. For this step, we prompt the LLM to classify the allegations in three classes: \textit{inconclusivo} (inconclusive), \textit{afastada} (not grounded in law), and \textit{configurada} (grounded in law). We provide the model with the allegations and the classification of each allegation obtained in the first step. We then ask the model to produce a decision about the \textit{Fumus boni iuris} of the case. The model uses the classification of the allegations to produce the decision. Before the decision is made, the model is instructed to produce a paragraph explaining the decision. The paragraph should explain the decision and the reasons for the decision. The paragraph is included in the initial instruction document.

\subsection{Recommendations} \label{sec:recommendations}

The INACIA pipeline concludes by generating a paragraph that, in this context, is called recommendations (i.e., the adjudication suggestions). The recommendations are a set of directions for the decision-making authority to address the case. Since the auditor assigned the case bears full responsibility for the adjudication recommendations, INACIA only suggests directions for the auditor to consider but the auditor has the final say on whether they accept the suggestion.
This pipeline task is relatively straightforward and recasts the other outputs from each task to construct a cohesive and coherent instruction document to be sent over to the panel of judges who will issue the final ruling on a case. To write adjudication recommendations, INACIA uses the following components:
\begin{itemize}
    \item The output from the admissibility examination, i.e., whether the case is admissible and the explanation of such classification.
    \item The outcome of the \textit{periculum in mora} and \textit{fumus boni iuris} analyses, i.e., whether there is any danger in delay and the case is likely to be successful based on its merits.
    \item The analysis of the case's merits, with the details on claims, their relationship with the jurisprudence and statutory law, and recommended courses of action.
\end{itemize}

Each of these inputs is then processed by the LLM, which had been previously instructed with guidelines and examples of recommendations authored by experienced auditors. The LLM generates paragraphs for each recommendations section, ensuring the final text is coherent and accurately reflects the proposals and analyses. The end product is a structured document that is broken down into several sections, each on different aspects of the case, such as the case information, decisions on admissibility, merits, etc.

This modular construction of the final instruction document makes sense both from the legal and the computational sense. From the legal perspective, it allows human reviewers to parse out the document into smaller chunks and use/correct them as they see fit. From the computational side, it means we do not have to reprocess everything in later stages if we need a small correction in one of the earlier stages of INACIA. This approach provides efficiency, especially in cases where, despite a potential flaw in one section, the rest of the information remains valid and useful.

The recommendations task is thus a critical element of INACIA, marking the transition from analysis to actionable insights and ensuring that the outcomes of the case review process are communicated with clarity and precision, consistent with the standards upheld by the TCU.

\section{Experiments} \label{sec:experiments}

This section describes the experiments we conducted to evaluate the INACIA pipeline. Although INACIA has a variety of project tasks, we only evaluated the recommendations task, which represents the system's final and most critical output. We first describe the dataset we used in the evaluation and then describe the evaluation techniques we used. Finally, we present the results of the experiments.

\subsection{Dataset Creation and Curation}

A validation dataset was created to assess the quality of the analyses performed by INACIA. This dataset consists of fully annotated cases for all information pieces extracted at each step of the pipeline. In total, we annotated 122 cases and 33 variables per case.\footnote{
The selection of these cases followed criteria set up by the research team and TCU to better evaluate INACIA given the wide variety of cases processed by the court: we blocked cases whose claims were violations of one procurement law, between 2019 and 2022, and which contained \textit{periculum in mora} analyses (such that we could recreate the complete analysis needed by TCU).
} Though the final evaluation is only performed with the recommendations task, we tracked all other variables for intermediate outputs and for adjusting each INACIA module to output the necessary results for the following module. The annotation of these cases consisted primarily of having the language model extract each variable from the initial instruction of every case. Since we only worked with disposed cases, we had the adjudication recommendation document (i.e., the initial instruction) available to us when creating this validation dataset, which served then as the single source of validation information. The split of extracted variables is (cf. Appendix \ref{sec:variables_validation_dataset}):

\begin{itemize}
    \item 23 variables in the basic information module;
    \item 5 variables in the admissibility examination module;
    \item 2 variables in the \textit{periculum in mora} module;
    \item 1 variable in the \textit{fumus boni iuris} module;
    \item 1 variable in the allegations and requests module;
    \item 1 variable in the recommendations module.
\end{itemize}

We instructed GPT-4 to extract all the information above from the initial instruction of each case to retrieve the labels INACIA would have to predict based on case filing documents (i.e. the \textit{representação} and supporting documents).\footnote{This process is exactly what TCU auditors do: from a set of documents filed by the plaintiff( i.e., the \textit{representação} and supporting documents) they draft the initial instruction.}
To guarantee that the validation dataset was being constructed correctly, we assigned one senior researcher in our team (Ph.D. level) to manually review 30 randomly sampled cases for all variables to then compare the researcher's data extraction to that of GPT. GPT achieved 100\% accuracy on this task, and we confidently proceeded to ask it to code the 122 selected cases.

Although we collect and validate a wide range of information, including basic case details, claims, admissibility assessments, and so on, the results do not carry equal significance. We emphasize the evaluation of the recommendations, as it is the last and most important step of the instruction process.


\subsection{Evaluation} \label{evaluation}

For evaluating the INACIA main output, we used the validation dataset described in the previous section. The evaluation aims to compare the outcomes produced by the INACIA pipeline with the expected results previously produced by human auditors. We only evaluated INACIA on the recommendations task (cf. Section \ref{sec:recommendations}), which is the consolidation of outputs of all intermediate tasks (listed in Sections \ref{sec:basic_info_extraction} to \ref{sec:plausability}) to write the initial instruction document.

The evaluation of recommendations could be done using term-matching metrics, such as BLEU, ROUGE, etc. However, 
evaluating free-form text generated by LLMs is a challenging task. While BLEU and ROUGE are commonly used to evaluate text generation tasks, they have shown low correlation with human judgment \cite{fabri2021summeval}. Approaches like GPTScore \cite{fu2023gptscore}, and G-eval \cite{liu2023geval} have been proposed to address this issue by using LLMs to evaluate the generated text. They have been shown to correlate more with human judgment than traditional metrics when evaluating text summarization.
However, these approaches are also unsuitable for evaluating the text generated by the INACIA pipeline because the generated text is not a summary of a document. Instead, the generated text is similar to what the auditor's adjudication recommendations were.

\begin{figure}[t]
  \centering 
  \includegraphics[width=8cm]{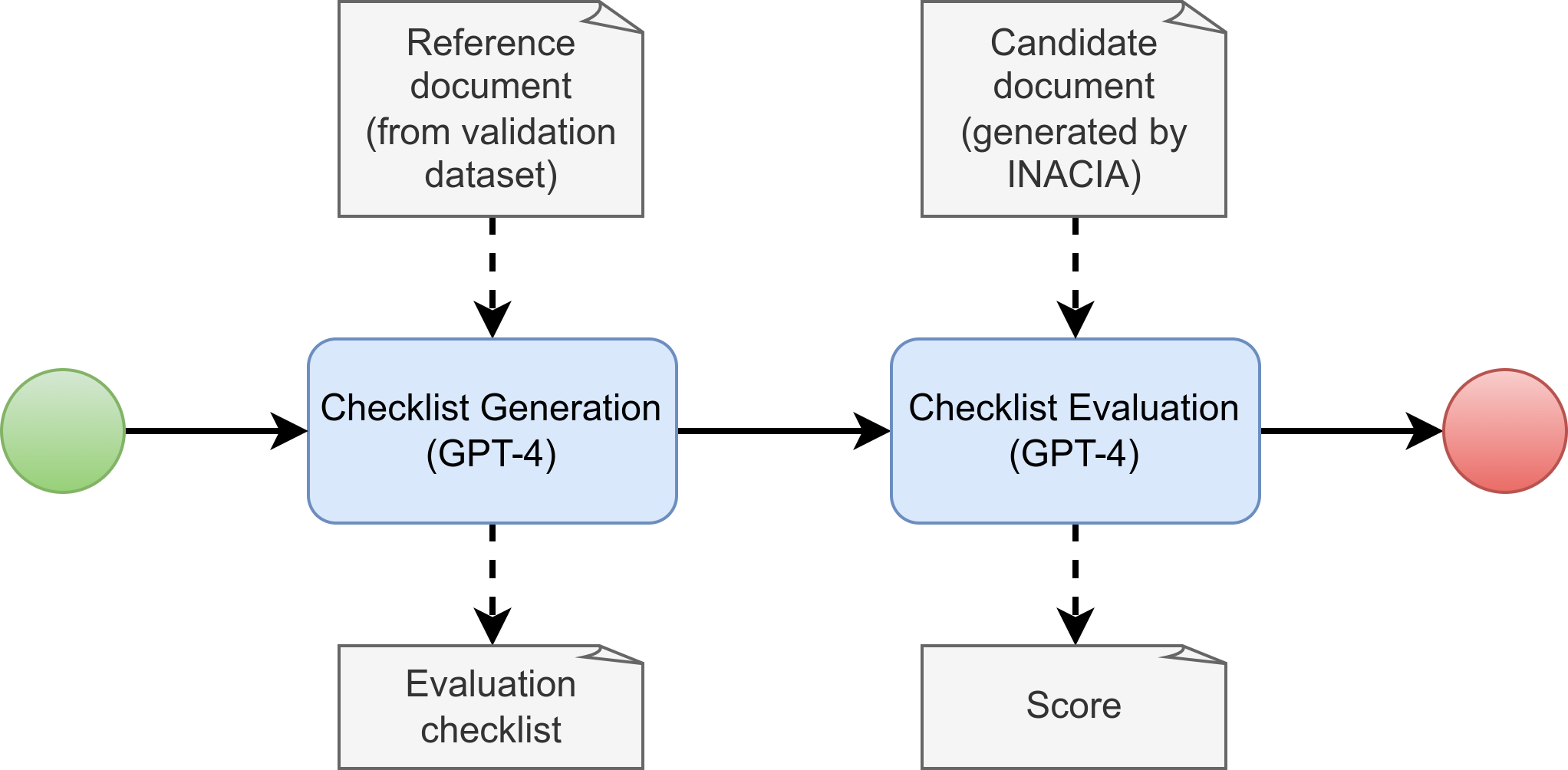}
  \caption{The checklist approach used to evaluate the recommendations.}
  \label{fig:checklist_pipeline}
\end{figure}

The objective of this evaluation thus is to assess whether the recommendations produced by INACIA are consistent with those produced by a human auditor. Intuitively, we want to evaluate if the candidate text (INACIA) contains the same elements as the reference text (human auditor). For this, we used the approach depicted in Figure \ref{fig:checklist_pipeline}. First, we prompted OpenAI's GPT-4 with the reference recommendations from the validation dataset, asking it to compose a checklist of the elements that should be present in the recommendations (i.e., the Checklist Generation step). The result is a list of elements that should be present in the recommendations (i.e., the evaluation checklist). Each element is a sentence with a boolean statement (e.g., \textquote{The case is admissible}).
We then prompted the model with the candidate recommendations alongside the evaluation checklist and asked it to evaluate if the text contained the checklist elements (i.e., the Checklist Evaluation step). Then, we can count the number of checklist elements present in the candidate recommendations and use this number to evaluate the quality of the extracted information. This produces a score that indicates how consistent the candidate recommendations are with the reference recommendations.
For example, if the checklist contains ten elements and the candidate recommendations contain eight elements, we can say that the candidate recommendations are 80\% consistent with the reference recommendations.

Figure \ref{fig:checklist_example} shows an example of a checklist of elements that should be present in the recommendations of an example case. It was composed by OpenAI's GPT-4 originally in Brazilian Portuguese and then translated to English for this paper. The checklist contains five elements that should be present in the candidate text. This way, we can evaluate the quality of the recommendations by comparing the checklist elements with the candidate recommendations. It is worth noting that the number of elements is not fixed and can vary from case to case.

\begin{figure}[t]
    \centering
    \footnotesize
\begin{subfigure}[t]{0.49\textwidth}
\begin{framed}
\begin{enumerate}[leftmargin=\parindent,align=left,labelwidth=\parindent,labelsep=1pt]
    \item Verificar se a representação atende aos requisitos de admissibilidade conforme Lei 8.666/1993, art. 113, § 1º, e os arts. 235 e 237, VII, do Regimento Interno do TCU, e o art. 103, § 1º, da Resolução - TCU 259/2014.\\
    \item Avaliar a possibilidade de apensar o processo atual ao TC 045.783/2021-7, com base no art. 36 da Resolução-TCU 259/2014, devido à conexão entre seus objetos.\\
    \item Decidir sobre o indeferimento da medida cautelar solicitada.\\
    \item Assegurar a comunicação ao representante sobre a decisão que será tomada.\\
    \item Considerar o arquivamento dos autos atuais, conforme o art. 169, I, do Regimento Interno do TCU.\\
\end{enumerate}
\end{framed}
\caption{Brazilian Portuguese version.}
\end{subfigure}
\begin{subfigure}[t]{0.49\textwidth}
\begin{framed}
    \begin{enumerate}[leftmargin=\parindent,align=left,labelwidth=\parindent,labelsep=1pt]
        \item Check whether the representation meets the admissibility requirements in accordance with Law 8,666/1993, art. 113, § 1, and arts. 235 and 237, VII, of the TCU Internal Regulations, and art. 103, § 1, of Resolution - TCU 259/2014.\\
        \item Evaluate the possibility of attaching the current process to TC 045.783/2021-7, based on art. 36 of Resolution-TCU 259/2014, due to the connection between its objects.\\
        \item Decide whether to reject the requested precautionary measure.\\
        \item Ensure communication to the representative about the decision that will be made.\\
        \item Consider archiving current records, in accordance with art. 169, I, of the TCU Internal Regulations.\\
    \end{enumerate}
\end{framed}
\caption{English version.}
\end{subfigure}
    \caption{Example of a checklist of elements that should be present in the recommendations. The checklist was composed by OpenAI's GPT-4 originally in Brazilian Portuguese and then translated to English for this paper.}
    \label{fig:checklist_example}
\end{figure}{}

We also used the candidate document to prompt the model to compose a checklist of the elements that should be present in the candidate document. We then prompted the model with the reference document and asked it to evaluate if it contained the checklist elements. Then, we can count the number of checklist elements present in the reference document and use this number to evaluate the quality of the extracted information. This way, we have two scores: the precision, which indicates how consistent the candidate document is with the reference document, and the recall, which indicates how consistent the reference document is with the candidate document. We then use the F1 score to evaluate the quality of the extracted information. The F1 score is the harmonic mean of the precision and recall. 

This approach has shown to have a good performance in assessing the quality of the extracted text in the TCU context. To evaluate it, we asked an auditor to analyze five examples in which the score is too low (less than 0.3) and too high (more than 0.7). The auditor concluded that the examples with low scores were not consistent with the reference text, and those with high scores were consistent with the reference text. This shows that the approach is suitable for evaluating the quality of the extracted text in the TCU context. Yet, we evaluated this approach on assessing the consistency of the automatically generated summary so we could compare it with GPTScore and G-eval. We used the SummEval Dataset \cite{fabri2021summeval} as a benchmark. The results shown in Table \ref{tab:check_eval} demonstrate that our approach correlates more with human judgment than GPTScore and G-eval on assessing how consistent a summary is concerning its original text. It also has a higher correlation with human judgment, considering the overall quality of the summary. This shows that our approach is suitable for evaluating the quality of the automatically generated text.

\begin{table}[t]

\caption{Check-eval results. Summary-level ($\rho$) Spearman and ($\tau$) Kendall-Tau correlations.}
\label{tab:check_eval}
\begin{tabular}{@{}lcccccccccc@{}}
\toprule
Method     & \multicolumn{2}{c}{Consistency} & \multicolumn{2}{c}{Relevance} & \multicolumn{2}{c}{Coherence} & \multicolumn{2}{c}{Fluency} & \multicolumn{2}{c}{AVG} \\ 
           & $\rho$        & $\tau$       & $\rho$       & $\tau$      & $\rho$       & $\tau$      & $\rho$      & $\tau$     & $\rho$    & $\tau$   \\ \midrule
G-eval     & 0.507           & 0.425         & \textbf{0.547}          & \textbf{0.433}        & \textbf{0.582}          & \textbf{0.457}        & 0.455         & 0.378       & 0.514       & 0.418     \\
GPTScore   & 0.449           &               & 0.381          &              & 0.434          &              & 0.403         &             & 0.417       &           \\
check-eval & \textbf{0.605}           & \textbf{0.570}         & 0.5028         & 0.4202       & 0.563          & 0.461        & \textbf{0.490}         & \textbf{0.446}       & \textbf{0.623}       & \textbf{0.493}     \\ \bottomrule
\end{tabular}
\end{table}

\section{Results} \label{sec:results}



Table \ref{tab:check_list_recommendations} presents the results of evaluating the recommendations generated by the INACIA pipeline using the checklist approach. This method assesses the alignment of the generated recommendations with the checklist of elements derived from the reference text. The table shows the mean, standard deviation (Std), minimum (Min), and maximum (Max) values for precision, recall, and the F1 score. These metrics provide insights into the consistency and completeness of the generated recommendations compared to the reference text. The average precision score of 0.592 indicates that, on average, 59.2\% of the checklist elements were found in the generated recommendations. This suggests a moderate level of accuracy in capturing the necessary elements in the recommendations. The average recall score is 0.401, indicating that roughly 40.1\% of the reference text elements were present in the generated recommendations. This lower recall suggests that the generated recommendations do not capture significant reference text elements. The average F1 score is 0.429 which reflects a balance between precision and recall but suggests room for improvement in both aspects.
The standard deviation values for precision (0.212), recall (0.275), and F1 (0.225) show a certain degree of variability in the performance across different cases. Overall, these results suggest that while there are instances where the recommendations closely match the reference text, there is a noticeable variation in the quality of the generated recommendations. The relatively lower recall scores indicate a need for improvement in capturing all the necessary elements from the reference text in the generated recommendations.

\begin{table}[b]
\caption{Results of the recommendations using the checklist approach.} 
\label{tab:check_list_recommendations}
\begin{tabular}{@{}lccc@{}}
\toprule
     & Precision & Recall & F1    \\ \midrule
Mean & 0.592 & 0.401 & 0.429 \\
Std & 0.212 & 0.275 & 0.225 \\
Min & 0.000 & 0.059 & 0.000 \\
Max & 1.000 & 1.000 & 0.933 \\ \bottomrule
\end{tabular}
\end{table}

Although the INACIA pipeline does not always produce the expected results, it is possible to follow the reasoning of the INACIA pipeline to understand the decisions made by the agent. This helps the auditor to find the evidence. For example, the INACIA pipeline may not classify the case as admissible regarding some criterion in the admissibility examination. However, it is possible to follow the reasoning of the INACIA pipeline to understand why the case was not classified as admissible. 
Figure \ref{fig:admissibility_examination_reasoning} shows an example of the reasoning produced by INACIA for the admissibility examination considering the public interest criterion. The reasoning explains why the case was classified as admissible regarding the public interest criterion. This helps the auditor to understand the decision made by INACIA and find the evidence that supports it.

\begin{figure}[t]
    \centering
    \footnotesize
\begin{subfigure}[t]{0.48\textwidth}
\begin{framed}
Explicação: As alegações do autor da representação indicam a existência de possíveis irregularidades no Pregão Eletrônico nº 03/2021 promovido pelo IBAMA. Entre as irregularidades apontadas, estão a não observância das exigências do instrumento convocatório e das leis vigentes da licitação pública, atos praticados por servidores do órgão que podem ser considerados obscuros e que ferem o caráter competitivo do certame, e a habilitação de uma empresa que não apresentou a certidão de falência exigida pelo edital.\\
\\
Caso essas alegações sejam confirmadas, elas podem indicar a existência de interesse público, conforme definido no Manual de Representações e Denúncias do TCU. Isso porque, se comprovadas, essas irregularidades podem representar um potencial risco de dano ao erário, configurar uma restrição indevida à competição, representar um potencial risco de inexecução ou execução insatisfatória do contrato, configurar uma fraude à licitação ou ao contrato, e representar um potencial risco de ineficácia na execução de política pública.\\
\\
Admissível: Sim
\end{framed}
\caption{Brazilian Portuguese version.}
\end{subfigure}
\begin{subfigure}[t]{0.48\textwidth}
\begin{framed}
Explanation: The allegations made by the author of the representation indicate the existence of possible irregularities in Electronic Auction No. 03/2021 promoted by IBAMA. Among the irregularities highlighted are non-compliance with the requirements of the call for proposals and the current laws governing public bidding, acts carried out by the agency's employees that may be considered obscure and that harm the competitive nature of the event, and the qualification of a company that does not present the bankruptcy certificate required by the notice.\\
\\
If these allegations are confirmed, they may indicate the existence of public interest, as defined in the TCU Representations and Complaints Manual. This is because, if proven, these irregularities may represent a potential risk of damage to the treasury, constitute an undue restriction on competition, represent a potential risk of non-execution or unsatisfactory execution of the contract, constitute fraud in the bid or the contract, and represent a potential risk of ineffectiveness in the execution of public policy.\\
\\\\
Admissible: Yes
\end{framed}
\caption{English version.}
\end{subfigure}
\caption{Example of the reasoning produced by INACIA for the admissibility examination considering the public interest criterion.}
\label{fig:admissibility_examination_reasoning}
\end{figure}{}

\section{Discussion} \label{sec:discussion}

\subsection{Limitations}

INACIA departs from common, simple uses of large language models in a few ways: \begin{enumerate*}
    \item it uses long context windows (and retrieval-augmented generation) techniques to extract information spread across multiple case documents;
    \item it requires the LLM to reason through many, often conflicting documents and identify the most important claims being made and present those to its user (e.g., evidence presented by plaintiff vs. jurisprudence);
    \item it allows the LLM to search for external documents that are relevant to each analysis before generating text for the user;
    \item it produces a list of adjudication recommendations with little knowledge of what information takes precedence (e.g., is it the law or jurisprudence, is it a more recent or less recent piece of legislation, is it one claim or all claims, etc.);
    \item its main product is free-form text with recommendations that are not easily compared to purported ground truth retrieved from work done by TCU auditors.
\end{enumerate*}
Naturally, thus, the system has limitations due to the inability of LLMs to perform all the tasks above. Besides these inherent limitations, there are practical limitations akin to large projects such as INACIA; e.g., the time constraints on developing each module and then integrating the whole application; the balance between developing a very advanced, cutting-edge but also unstable technology; the gap in business, context-specific knowledge between the research team and TCU auditors, etc. Despite these limitations, we firmly believe INACIA has great potential to assist TCU staff in their tasks and, therefore, add a lot of value to the TCU case processing structure.

\subsection{Ethical Implications}

Whenever artificial intelligence systems are being developed in real-world applications, one of the first concerns of decision-makers responsible for the system's adoption is whether the models are biased or fair. While acknowledging these as real issues in any AI system, we believe INACIA to be relatively shielded against such concerns due to the nature of the case processing steps and information provided to the language models to complete each step in the INACIA pipeline. First and foremost, the complaints INACIA processes are not filed against individual defendants nor any individual trait is used in the analysis of case components. Secondly, the techniques used to prompt INACIA are all grounding its answers onto information belonging to TCU's knowledge base, such as jurisprudence, legal statutes and norms, and previous cases handled by the audit court. This structure ensures that INACIA's adjudication recommendations are based on established legal principles and precedents, significantly mitigating the risks of bias.

Another important ethical consideration is the transparency, explainability, and accountability behind INACIA's decision-making process. Since the potential consequences of INACIA's recommendations can be large, it is imperative that the reasoning behind these decisions is clear and understandable. We believe this risk to be limited since none of the decision factors used in INACIA prompts contains elements departing from current adjudication practices at TCU. TCU staff has provided extensive support and auditing materials so that INACIA could replicate as close as possible the work of a TCU auditor. Secondly, TCU auditors and eventually the court's judicial panel bear ultimate responsibility for any case decision. TCU has not removed, nor plans on removing, any human oversight of INACIA's outputs.\footnote{
There are other relatively minor risks that any new system carries, such as data security and privacy and job displacement, that are not particular to INACIA. As far as data security and privacy go, INACIA is being built to be compatible with existing TCU systems without adding any new risk. As to potential job displacement risks, TCU is precisely looking to replace tasks in its staff's work assignments with AI such that staff can shift their skills to other, more important tasks within TCU.
}

\section{Conclusions} \label{sec:conclusions}

In this paper, we presented INACIA, an innovative system designed to integrate Large Language Models (LLMs) into the Brazilian Federal Court of Accounts operational framework. Our experiments and analyses reveal that INACIA holds significant potential to support the auditor in the audit process with greater speed and security in analyzing cases and making decisions.

The utilization of LLMs for processing and analyzing case documents demonstrated a moderate level of accuracy in capturing essential elements and information from cases. The system's ability to handle various stages of case processing, from basic information extraction to admissibility examination, precautionary measure analysis, legal plausibility assessment, and final recommendations generation, showcases its versatility and adaptability to the complex requirements of legal proceedings.

Moving forward, we will focus on enhancing INACIA's capabilities through continuous development and improvements. Besides, we will explore implementing an auditing system based on INACIA's architecture, allowing us to evaluate the system's performance in a real-world setting. In conclusion, INACIA represents a significant step forward in applying artificial intelligence in the legal domain. Its development and implementation in the Brazilian audit courts can serve as a blueprint for similar initiatives globally. 


\bibliographystyle{ACM-Reference-Format}
\bibliography{sample-base}
\pagebreak

\appendix

\section{List of Variables in the Validation Dataset} \label{sec:variables_validation_dataset}


\footnotesize

\begin{longtable}{|p{5cm}|p{9cm}|} 
\caption{List of variables in the validation dataset.}\label{tab:variable_List}\\
\hline  
\textbf{Variable Name} & \textbf{Description} \\  
\hline  
\endhead 
\multicolumn{2}{r}{Basic Information Extraction} \\
\hline
numero\_processo & Individual case ID \\  
info\_uasg & Federal agency under investigation (unique ID) \\  
info\_author\_name & Plaintiff\'s name \\  
info\_author\_doc & Plaintiff\'s document ID \\  
info\_author\_doc\_type & Plaintiff\'s document type \\  
info\_author\_doc\_complement\_type & Plaintiff's additional document type information \\  
info\_author\_doc\_complement & Plaintiff's additional document information \\  
info\_case\_summary & Case summary \\  
info\_case\_jurisdictional\_unit & Federal agency under investigation (name) \\  
info\_case\_subject & Case subject \\  
info\_case\_dispute\_validity & Period during which procurement call remained open for proposals \\  
info\_case\_type & TCU case classification \\  
info\_case\_dispute\_mode & Procurement awarding mechanism \\  
info\_case\_dispute\_criterion & Procurement awarding mechanism criteria \\  
info\_case\_dispute\_value & Procurement contract value \\  
info\_case\_dispute\_value\_type & Procurement contract value estimated or observed \\  
info\_case\_status & Procurement status \\  
info\_case\_status\_ambiguous & Procurement status ambiguous \\  
info\_procurement\_id & Procurement ID \\  
info\_procurement\_phase & Procurement stage \\  
info\_procurement\_type & Procurement type \\  
info\_procurement\_type\_id & Procurement type ID \\  
info\_procurement\_law & Law applicable to procurement call \\  
\hline
\multicolumn{2}{r}{\textit{Periculum in mora}} \\
\hline
cautelar\_perigo\_demora & Periculum in mora present (per TCU auditor evaluation) \\  
cautelar\_perigo\_demora\_rev & Reverse periculum in mora present (per TCU auditor evaluation) \\  
\hline
\multicolumn{2}{r}{\textit{Fumus boni iuris}} \\
\hline
cautelar\_plausibilidade & Fumus boni iuris (per TCU auditor evaluation)  \\
\hline
\multicolumn{2}{r}{Admissibility Examination} \\
\hline
admiss\_legitimidade & Legitimacy confirmation \\  
admiss\_linguagem & Clear writing and language confirmation \\  
admiss\_indicio & Existence of evidence confirmation \\  
admiss\_competencia & Competency confirmation \\  
admiss\_interesse\_pub & Existence of public interest \\  
\hline
\multicolumn{2}{r}{Allegations and Requests} \\
\hline
alegacoes\_txt & List of allegations \\  
\hline  
\multicolumn{2}{r}{Recommendations} \\
\hline
encaminhamentos\_txt & List of recommendations made by TCU auditor \\
\hline
\end{longtable}  

\end{document}